\documentclass{article} % For LaTeX2e
\usepackage{iclr2026_conference,times}
\iclrfinalcopy
\usepackage{amsmath,amsfonts,bm}

\def\eqref#1{equation~\ref{#1}}
\def\1{\bm{1}}

\DeclareMathAlphabet{\mathsfit}{\encodingdefault}{\sfdefault}{m}{sl}
\SetMathAlphabet{\mathsfit}{bold}{\encodingdefault}{\sfdefault}{bx}{n}

\usepackage{hyperref}

\usepackage{url}            % simple URL typesetting
\usepackage{booktabs}       % professional-quality tables
\usepackage{amsfonts}       % blackboard math symbols
\usepackage{nicefrac}       % compact symbols for 1/2, etc.
\usepackage{microtype}      % microtypography        % colors
\usepackage{amsmath}        % for \text in math mode
\usepackage{multirow}
\usepackage{makecell}
\usepackage{listings}
\usepackage{ulem}
\usepackage{subcaption}
\usepackage{pifont}
\usepackage{colortbl}

\usepackage{tcolorbox}
\tcbuselibrary{listings, breakable} % 必须加载 listings 和 breakable 库
\usepackage{listings}

\title{Fine-tuning Large Language Model for Automated Algorithm Design}

\author{
Fei Liu\textsuperscript{1}, 
Rui Zhang\textsuperscript{1}, 
Xi Lin\textsuperscript{2}, 
Zhichao Lu\textsuperscript{1} \& 
Qingfu Zhang\textsuperscript{1} \\
\textsuperscript{1} City University of Hong Kong, Hong Kong, China\\
\textsuperscript{2} Xi'an Jiaotong University, Xi'an, China \\
\texttt{\{fliu36-c, rui.zhang.cs\}@my.cityu.edu.hk}, \texttt{xi.lin@xjtu.edu.cn}, \\ \texttt{\{zhichaolu, qingfu.zhang\}@cityu.edu.hk}
}

\begin{document}

\maketitle
\begin{abstract}

The integration of large language models (LLMs) into automated algorithm design has shown promising potential. A prevalent approach embeds LLMs within search routines to iteratively generate and refine candidate algorithms. However, most existing methods rely on off-the-shelf LLMs trained for general coding tasks, leaving a key question open: \textit{Do we need LLMs specifically tailored for algorithm design?} If so, how can such LLMs be effectively obtained and how well can they generalize across different algorithm design tasks? In this paper, we take a preliminary step toward answering these questions by exploring fine-tuning of LLMs for algorithm design. We introduce a \textit{Diversity-Aware Rank-based (DAR)} sampling strategy to balance training data diversity and quality, then we leverage direct preference optimization to efficiently align LLM outputs with task objectives. Our experiments are primarily conducted on \textit{Llama-3.2-1B-Instruct} and \textit{Llama-3.1-8B-Instruct} across three distinct algorithm design tasks, with \textit{openPangu-Embedded} models additionally included as auxiliary comparisons on the admissible set problem. Results suggest that fine-tuned LLMs can significantly outperform their off-the-shelf counterparts with the smaller \textit{Llama-3.2-1B-Instruct} and match the larger \textit{Llama-3.1-8B-Instruct} on the admissible set problem. Moreover, we observe promising generalization: LLMs fine-tuned on specific algorithm design tasks also improve performance on related tasks with varying settings. These findings highlight the value of task-specific adaptation for LLMs in algorithm design and open new avenues for future research. Our code is publicly available at \url{https://github.com/RayZhhh/dpo-aad}.

\end{abstract}

\section{Introduction}
The emerging field of automated algorithm design (AAD) with large language models (LLMs) has attracted growing attention for its potential to automate the synthesis of expert-level algorithms~\citep{liu2024systematic, liu2024evolution, romera2024mathematical}. 
A prevailing paradigm in this space combines LLMs within search strategies, where the LLM focuses on generating candidate algorithms and the search procedures control the quality and refinement of these algorithms in an iterative manner~\citep{zhang2024understanding}.
% candidate algorithms are generated, evaluated, and refined  
% the search strategy iteratively performs the trial-and-error operations, and the LLM serves as the engine to drive the search. 
This framework has led to notable advances across a spectrum of algorithmic development tasks, including combinatorial optimization~\citep{liu2024evolution, ye2024reevo}, Bayesian optimization~\citep{yao2024evolve}, and black-box optimization~\citep{van2024llamea}, to name a few.  

% \fei{
% 尽管目前已经取得了成果，但是目前的方法都普遍存在共同的问题：1）需要requests很多次大模型，2）似乎不同的大模型类型对结果的影响不大或者没有明确的优劣。似乎提示我们需要一个specialized AAD.
% }

Despite these preliminary successes, most existing LLM-driven AAD approaches rely on off-the-shelf LLMs, posing two limitations: \ding{182} they require a large number of queries to LLMs, resulting in substantial computational overhead~\citep{romera2024mathematical, novikov2025alphaevolve}, and \ding{183} these methods exhibit marginal performance variations across different choice of LLMs~\citep{liu2024llm4ad, zhang2024understanding}, suggesting that current LLMs may lack inductive biases suited to algorithm design. 
These observations raise a potential need for specialized LLMs explicitly trained for algorithm design tasks. 
%
% \vspace*{\fill} 
% \begin{quote} 
% \centering 
% \textit{Do we need LLMs specially tailored for algorithm design?}
% \end{quote}
% \vspace*{\fill}
% strongly indicate the need for a specialized LLM explicitly trained for automated algorithm design tasks. 
While prior work has explored domain-specific LLMs for general coding (e.g., programming) tasks \citep{jiang2024survey} and optimization problem formulation \citep{huang2025orlm}, and some recent efforts have fine-tuned LLMs during the search process to improve AAD performance~\citep{huang2025calm, surina2025algorithm}, the development of LLMs tailored specifically for automated algorithm design remains largely underexplored. 
% On one hand, existing work has explored training domain-specific LLMs for general coding tasks \citep{jiang2024survey} and optimization problem formulation \citep{huang2025orlm};
% %
% on the other hand, a couple of recent efforts have tried to improve AAD performance by fine-tuning the LLMs during the search process~\citep{huang2025calm, surina2025algorithm}.
% %
% Nevertheless, developing LLMs explicitly tailored for automated algorithm design remains largely unexplored. 
% have investigated fine-tuning LLMs during iterative algorithm searches to improve performance on specific target tasks~\citep{huang2025calm, surina2025algorithm}

%
% Recent efforts have investigated fine-tuning LLMs during iterative algorithm searches to improve performance on specific target tasks~\citep{huang2025calm, surina2025algorithm}, but fundamental questions remain unanswered: How should domain-specific algorithm design LLMs be trained, and how effective are these models across diverse tasks?

Moreover, fine-tuning LLMs for algorithm design poses unique challenges distinct from conventional code generation~\citep{jiang2024survey} or mathematical reasoning~\citep{ahn2024large} tasks.
Algorithm design tasks rarely have clear ground-truth labels as optimal algorithms may not exist and cannot be evaluated by a single performance metric. The algorithm design process benefits from exploring diverse algorithms—even suboptimal ones—as they may introduce novel ideas or structural approaches that provide valuable insights and ultimately improve the final design. 
These characteristics render existing fine-tuning techniques insufficient for addressing the inherent complexity of fine-tuning LLMs for AAD.

In this work, we take a preliminary yet foundational step towards developing specialized LLMs for algorithm design tasks. Our investigation is guided by the following two research questions:
% \begin{itemize}
%     \item \textbf{Q1:} How to prepare training data for fine-tuning LLMs on algorithm design tasks?
%     \item \textbf{Q2:} Can a small fine-tuned LLM outperforms larger LLMs in algorithm design?
%     \item \textbf{Q3:} How about the generalization performance of fine-tuned LLMs across related algorithm design tasks?
% \end{itemize} 
\begin{itemize}
    \item \textbf{RQ1:} How can we effectively obtain LLMs specialized for algorithm design?
    \item \textbf{RQ2:} How well can these LLMs generalize across different algorithm design tasks? 
\end{itemize}

To address \textbf{RQ1}, we fine-tune general-purpose, open-source LLMs—specifically, \textit{Llama-3.2-1B-Instruct} and \textit{Llama-3.1-8B-Instruct}~\citep{grattafiori2024llama3herdmodels}—on algorithm design problems. For ASP, we additionally include OpenPangu models as auxiliary comparison models.
We introduce a Diversity-Aware Rank-based (DAR) Sampling strategy to construct diverse preference pairs, which serve as training data for fine-tuning via Direct Preference Optimization (DPO)~\citep{rafailov2023direct}. We evaluate the resulting LLMs against their original counterparts in two settings: (i) random sampling, and (ii) integration within an existing AAD framework.

To address \textbf{RQ2}, we assess the generalization capabilities of LLMs fine-tuned on the Capacitated Vehicle Routing Problem (CVRP) across two scenarios: 
(i) generalization to variant settings of the same problem (e.g., CVRP instances with different sizes and capacity constraints), 
and (ii) transfer to a related but distinct algorithm design task—namely, the Traveling Salesman Problem (TSP). 
These evaluations allow us to examine both in-distribution and out-of-distribution generalization performance.

Our \textbf{key findings} are summarized as follows.
\begin{itemize}
    \item  Fine-tuning LLMs specifically for algorithm design is necessary and feasible. The proposed Diversity-Aware Rank-based Sampling strategy enables effective and robust LLM fine-tuning, underscoring the importance of considering diversity in algorithm preference pair construction.
    
    \item Fine-tuned LLMs significantly improve their capabilities in (1) algorithm design with LLM-based random sampling (Sec.~\ref{sec:zero_shot_performance}), (2) algorithm design with LLM-based iterative search (Sec~\ref{sec:search_performance}), and (3) similar/related algorithm design tasks (Sec.~\ref{sec:generalized_performance}). Notably, \textit{Llama-3.2-1B-Instruct} trained with our method matches the performance of \textit{Llama-3.1-8B-Instruct}.
\end{itemize}

\section{Fine-tune LLM for AAD}

% \subsection{Preliminary}

% \fei{
% notations: \\
% $D$ represents database \\
% $a_i$ represents $i$-th algorithm \\
% $s_i$ represents $i$-th sample, each sample is a triple, $\{p,a^+,a^-\}$ \\

% }

\begin{figure}[t]~\label{framework}
    \centering
    \includegraphics[width=0.98\linewidth]{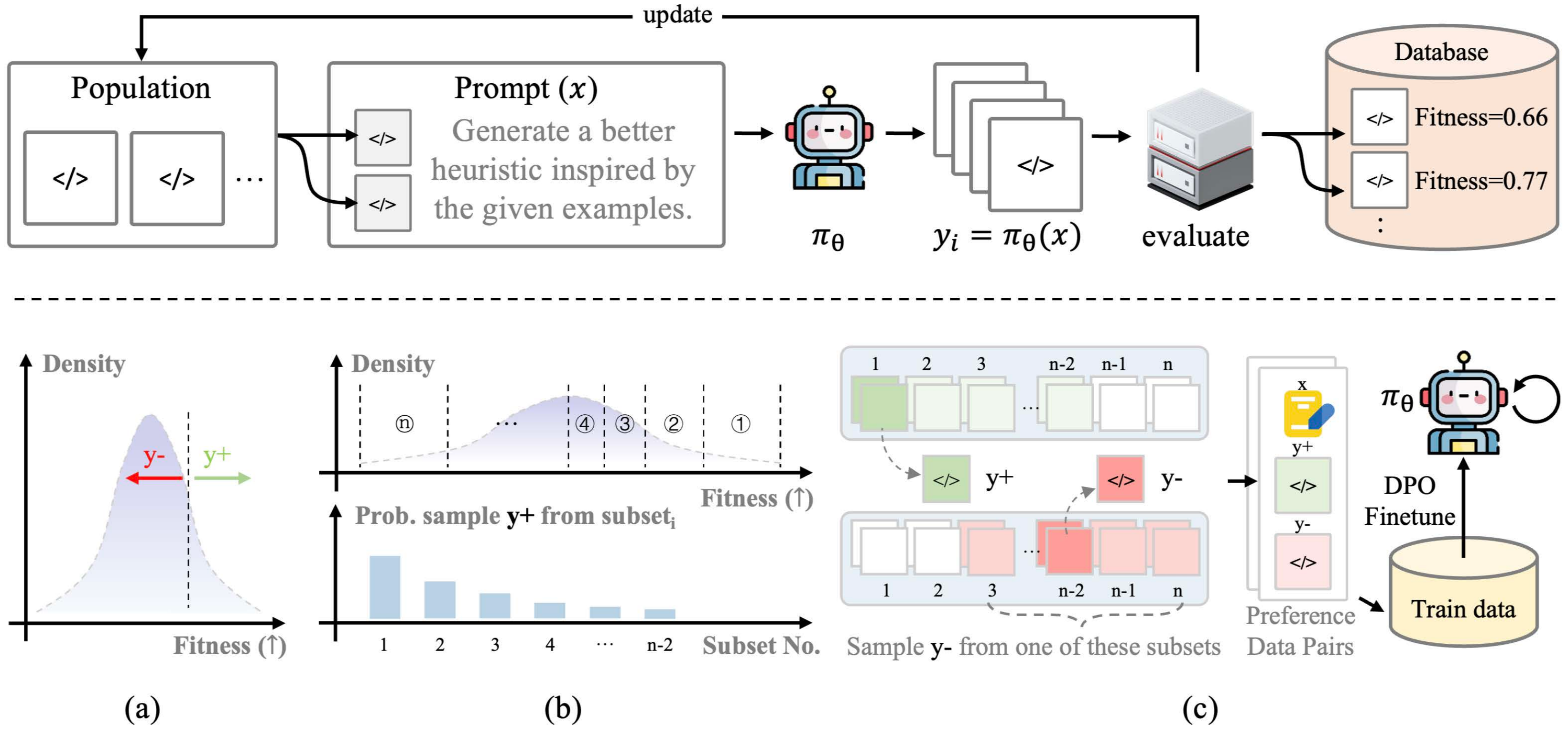}
    \caption{
    \textbf{Upper section:} LLM-based automated algorithm design methods iteratively optimize algorithms. Through this, algorithms and their fitness are preserved in the database $D$. The knowledge and experiences incorporated in the database subsequently improve the capabilities of the LLM. \textbf{Lower section:} (a) Traditional sampling relies on continuous fitness values and often suffers from unstable preference gaps. 
    (b) Our method discretizes the fitness space into ranked subsets, enabling structured sampling of high-quality $y_+$ and clearly worse $y_-$. 
    (c) These sampled pairs are combined with prompts and code templates to form training triples (samples) $(x, y_+, y_-)$ for fine-tuning.}
    \label{fig:offline}
\end{figure}

Rather than training algorithm design LLMs from scratch, we adopt a fine-tuning approach to adapt LLMs for automated algorithm design tasks. Among various learning methods, we employ Direct Preference Optimization (DPO), a reward-free method that trains models to prefer high-quality outputs over inferior ones using preference pairs. %Unlike other approaches (e.g., on-policy RL~\citep{guo2025deepseek} and supervised fine-tuning), DPO eliminates the need for reward models or heuristics while mitigating overfitting risks.

As shown in Figure~\ref{fig:offline}, our framework consists of two stages: 1) Data Generation: We use LLM-driven iterative algorithm search (e.g., EoH~\citep{liu2024evolution}) to generate diverse algorithms (\textbf{Upper section}). 2) Preference Learning: The collected algorithms are sampled to compose preference pairs (samples), enabling the LLM to learn preferred designs over less favoured ones (\textbf{Lower section}).
The fine-tuned LLMs can be used to generate algorithms.

\subsection{Data Generation}
We employ LLMs in iterative search methods (e.g., EoH~\citep{liu2024evolution} and FunSearch~\citep{romera2024mathematical}) to generate algorithms. These search methods maintain a population of algorithms, using LLMs to generate new algorithms or refine existing ones, thereby evolving the population over time. Unlike repeated sampling with LLMs, this approach produces high-quality, diverse algorithms~\citep{zhang2024understanding}, enabling more effective learning. Throughout this process, we record all valid algorithms generated to form the algorithm database $D$. These algorithms in $D$ are subsequently used to construct training datasets $D_t$ using a diversity-aware rank-based sampling strategy, as detailed in the next section.

\subsection{Diversity-Aware Rank-Based Sampling}

A natural and intuitive method for constructing preference pairs is to define a fitness threshold and sample positive algorithms ($y_+$) above this threshold and negative algorithms ($y_-$) from those below it (as illustrated in Figure \ref{fig:offline}(a)). 
A similar idea is also adopted in prior work, such as EvoTune \citep{surina2025algorithm}, where only those preference pairs in which the $y_+$ exhibits top-tier performance are retained for training.
While this strategy can ensure high-quality positive algorithms, discarding a large number of potentially informative samples with mid-range performance may limit the diversity of preference signals available to the learning model.

% While this strategy can ensure high-quality positive samples, only the top algorithms are selected, thereby discards a large number of potentially informative samples, limiting the diversity of training signals.

% two major limitations: (1) the resulting dataset tends to lack diversity, as only the top-performing algorithms are selected; and (2) the binary thresholding discards a large number of potentially informative samples with mid-range quality, thereby limiting the variety of training signals available.

To address these limitations, we propose a diversity-aware rank-based sampling strategy that constructs preference pairs in a more structured yet flexible way. This method aims to strike a balance between \textbf{quality emphasis} (favoring stronger candidates) and \textbf{diversity} (preserving a range of heuristic qualities), leading to more informative and robust preference supervision.

Let the algorithm database be denoted by $D = \{y_1, \dots, y_N\}$, where the candidates are sorted in \textbf{descending} order of fitness, i.e., $y_1$ is the best and $y_N$ the worst. We partition $D$ into $M$ \textit{equally-sized} and disjoint subsets:
$$
S_1,\; S_2,\; \dots,\; S_M, \quad \text{with } |S_m| = \left\lfloor \frac{N}{M} \right\rfloor \quad (m=1,\dots,M),
$$
so that $ S_1 $ contains the top-ranked, $ S_2 $ the next best, and so on (see Figure~\ref{fig:offline}(b)). We perform following steps to obtain a preference sample pair $ (x, y_+, y_-) $:

\textbf{1. Subset selection (biased toward higher quality).} From the first $ M{-}2 $ subsets, we pick an index $ i \in \{1, \dots, M{-}2\} $ with probability
$$
\Pr(i) = \frac{\exp((M-2-i)/\tau)}{\sum_{k=1}^{M-2} \exp(k/\tau)},
% \tag{1}
$$
where $ \tau > 0 $ is a \textit{temperature} hyperparameter (we set to $3.0$ by default). A smaller $ \tau $ sharpens the distribution, increasing the chance of drawing from higher-quality subsets. The \textit{temperature} $ \tau $ controls exploitation vs.\ exploration: When $ \tau \to 0 $ the positive sample will almost always draw from the very best subset(s); While when $\tau \to \infty $, it reduces to uniform sampling over the first $ M{-}2 $ subsets.

\textbf{2. Positive algorithm.} Sample one candidate uniformly from the chosen subset:
$$
y_+ \sim \text{Uniform}(S_i).
$$
    
\textbf{3. Negative algorithm.} To ensure a clear performance gap, we skip the nearest subset (i.e., $S_{i+1}$) and sample uniformly from the rest subsets:
$$
y_- \sim \text{Uniform}\left( \bigcup_{j=i+2}^{M} S_j \right).
$$
Note that we skip subset $ S_{i+1} $ in this step, which enforces a minimum gap of one quality tier, yielding clearer supervision signals.
    
\textbf{4. Preference sample construction.} A major distinction in our implementation of standard DPO lies in how we construct the prompt $x$. Different from standard DPO, where $y_+$ and $y_-$ algorithms are conditioned on a prompt $x$, 
%the $y_+$ and $y_-$ algorithms are conditioned on a prompt $x$. However, during the search phase, the $x$ typically comprises multiple few-shot algorithm examples. We observe that using such few-shot prompts directly as $x$ leads to instability in DPO fine-tuning. To address this,
we adopt a fixed prompt template consisting of two components: (1) a description of the algorithm design task to be solved, and (2) a function template and skeleton representing the expected format of the algorithm. %This design choice has two advantages. 
%First, using a fixed prompt format allows us to reuse historical algorithm data generated under different contexts. Second, a standardized and task-aware context provides consistent input to enhance the model’s generalization capability across different problem instances.
The resulting preference sample pair is a triplet $(x,\; y_+,\; y_-)$. Once a preference sample pair is obtained, we remove $y_+$ and $y_-$ from the database to eliminate duplication. This process is repeated to construct a training dataset $D_t$. 

Empirical studies in Sec. \ref{sec:expt1} demonstrate that the proposed sampling strategy maintains a balance between selecting high-quality heuristics and preserving diversity, thus producing rich and instructive training signals for algorithm preference learning.

% \begin{CJK*}{UTF8}{gbsn}
% \rui{这个地方要强调一下我们使用了统一的x}\fei{可以提一嘴？}
% \end{CJK*}

\subsection{DPO Fine-tuning}

We employ Direct Preference Optimization (DPO)~\citep{rafailov2023direct} to fine-tune LLMs using constructed training samples. The LLM acts as a policy $\pi_\theta(y|x)$, where $x$ is an input prompt and $y$ is a generated algorithm. Our objective is to optimize the LLM's preferences toward high-performing algorithms while maintaining generalization.

To optimize the policy $\pi_\theta$, we use an objective with regularization to ensure the outputs remain close to those of the reference model $\pi_{\text{ref}}(y|x)$:

\begin{align}
    \max_{\pi_\theta} \mathbb{E}_{x \sim \mathcal{D}_t, y \sim \pi_\theta}[r(x, y)] - \beta \mathbb{D}_{\text{KL}}[\pi_{\text{ref}}(\cdot|x) \parallel \pi_{\theta}(\cdot|x)],
\label{RL-objective}
\end{align}

where $\mathbb{D}_{\text{KL}}$ is the reverse KL-divergence, $\beta$ controls the regularization strength, and $\pi_{\text{ref}}$ is the initial base LLM policy $\pi_\theta^{0}$. While reward maximization methods like PPO~\citep{schulman2017proximal} exist, they are computationally expensive. Instead, we formulate the task as preference optimization, allowing algorithm ranking based on $r(x,y)$.

For a training dataset $\mathcal{D}_{\text{t}} = \{(x^i, y_{+}^{i}, y_{-}^{i})\}_{i=1}^{n}$, the loss function is:

% \begin{align}
% \mathcal{L}(\pi_{\theta}; \pi_{\text{ref}}) = \mathbb{E}_{(x, y_+, y_-) \sim \mathcal{D}_{\text{pref}}} \left[ -\log \sigma \left( \beta f'\left(\frac{\pi_\theta(y_+|x)}{\pi_{\text{ref}}(y_+|x)}\right) - \beta f'\left(\frac{\pi_\theta(y_-|x)}{\pi_{\text{ref}}(y_-|x)}\right) \right) \right],
% \label{DPO}
% \end{align}

\begin{align}
\mathcal{L}_{\text{DPO}} = -\mathbb{E}_{(x, y_+, y_-) \sim \mathcal{D}_{\text{t}}} \left[ 
\log \sigma\left( 
\beta \left( 
\log \frac{\pi_\theta(y_+|x)}{\pi_{\text{ref}}(y_+|x)} 
- \log \frac{\pi_\theta(y_-|x)}{\pi_{\text{ref}}(y_-|x)} 
\right) 
\right) 
\right],
\label{DPO}
\end{align}

where $\sigma$ is the sigmoid function. This approach eliminates the need for separate reward models or complex reward heuristics used in RL methods like GRPO~\citep{huang2025calm}, enabling efficient off-policy optimization.

% \begin{CJK*}{UTF8}{gbsn}
% \rui{这个地方要改下，没有reverse kl啥事儿了，然后最后说一下我们用了lora}
% \end{CJK*}

We apply low-rank adapters (LoRA)~\citep{hu2022lora} to fine-tune the model efficiently. LoRA introduces a small number of trainable parameters into existing layers of the model, allowing effective adaptation without updating the full parameter set. 
The adoption of LoRA is further motivated by the inherent challenges of our task: collecting sufficient algorithm data is expensive, as each candidate algorithm must undergo a computationally expensive evaluation phase during search. 
Therefore, applying LoRA is particularly important in our case, as the size of our training data may not be sufficient to support full fine-tuning without overfitting. Empirically, we have observed that full fine-tuning failed to converge during training on our datasets.

\section{Experimental Studies}
% \subsection{Experimental Settings}

\subsection{Algorithm Design Tasks}
% We demonstrate three algorithm design tasks:

% \item Online Bin Packing (OBP) aims to pack a collection of items using the fewest bins with fixed capacity $C$. In this work, we consider the online bin packing scenario where items are allocated to bins once they arrive. We consider the widely used Weibull~\cite{} dataset, with five instances where each comprises $5k$ items sampled from the Weibull distribution. The capacity of bins $C$ is set to $100$ as consistent with prior works~\cite{}.

\paragraph{Admissible Set Problem (ASP)}
ASP aims to maximize the size of the set while fulfilling the criteria below: (1) The elements of the set are vectors belonging to $\{0,1,2\}^n$. (2) Each vector has the same number $w$ of non-zero elements but a unique support. (3) For any three distinct vectors, there is a coordinate in which their three respective values are $\{0,1,2\}$, $\{0,1,2\}$, $\{0,1,2\}$. Following prior works~\citep{romera2024mathematical}, we set $n=15$ and $w=10$ in this work. 

\paragraph{Traveling Salesman Problem (TSP)}
TSP aims to find a route that minimizes the total traveling distance for a salesman required to visit each city exactly once before returning to the starting point. We investigate the constructive heuristic design for TSP. Specifically, we adopt an iteratively constructive framework to start from one node and iteratively select the next node until all nodes have been selected and back to the start node. The task is to design a heuristic for choosing the next node to minimize the route length.

\paragraph{Capacitated Vehicle Routing Problem (CVRP)}
CVRP aims to minimize the total traveling distances of a fleet of vehicles given a depot and a set of customers with coordinates and demands. The problem is constrained by: (1) The vehicles start from the depot and return to the depot; (2) Each customer should be visited only once; (3) All the demands should be satisfied while the capacity of the vehicle should not be exceeded. Similar to TSP, we adopt an iteratively constructive framework to start from one node and iteratively select the next node until all nodes have been selected and return to the depot. The task is to design a heuristic for selecting the next node to minimize the total route length with all constraints satisfied.

\subsection{Experimental Settings}
\paragraph{Data Generation} For each algorithm design task, we collect a diverse set of algorithmic solutions from existing results produced by FunSearch and EoH. These results are generated by \textit{Llama-3.1-8B-Instruct} \citep{grattafiori2024llama3herdmodels}. To standardize the code in the database, we first preprocess the raw code implementations by unifying their function templates, including consistent function names, docstrings, and input-output formats. Next, we discard identical algorithms by checking the code strings. Ultimately, we obtain approximately 60,000 unique algorithms for the ASP and CVRP problems, respectively. We adopt LLM4AD~\citep{liu2024llm4ad} implementations for both FunSearch and EoH.

\paragraph{Fine-tuning} We fine-tune each LLM for five epochs with a batch size of eight.
% with each batch consisting of a single instance processed across all GPUs. To accommodate the computational load, we employed a gradient accumulation strategy over two steps. The training regimen varied with the sample size: we conducted five epochs for 10,000 samples and extended this to fifty epochs for a smaller set of 1,000 samples, aiming to compensate for the reduced data exposure. 
We initiate the learning rate at 5e-6, applying a cosine decay schedule and a warmup rate of $0.05$ to reduce the rate over the training period gradually. The model processes inputs up to a maximum length of 2048 tokens. For DPO, we set the $\beta=0.4$ and utilize LoRA with settings of $r=64$ and $\alpha=32$, alongside a dropout rate of 0.05. The fine-tuning and inference processes for \textit{Llama-3.2-1B-Instruct} (Llama-1B) \citep{grattafiori2024llama3herdmodels} and \textit{Llama-3.1-8B-Instruct} (Llama-8B) \citep{grattafiori2024llama3herdmodels} are executed on NVIDIA L20 GPUs, while those for \textit{openPangu-Embedded-1b-v1.1} (Pangu-1B) \citep{chen2025panguembedded} and \textit{openPangu-Embedded-7b-v1.1} (Pangu-7B) \citep{chen2025panguembedded} are conducted on a Huawei Ascend 910B server. We use \texttt{trl} library~\citep{vonwerra2022trl} for DPO implementations and \texttt{vllm} library~\citep{kwon2023efficient} for efficient LLM inference.

\paragraph{Automated Algorithm Design}
We test the fine-tuned LLMs in two types of settings: 1) repeated sampling, where we use the same prompt to instruct LLMs to generate algorithms many times. This setting can show the effectiveness of fine-tuning in a straightforward way. 2) Iterative algorithm design, where we use the fine-tuned LLMs in EoH and FunSearch for automated algorithm design and compare to the results using pre-trained LLMs without fine-tuning. We set the maximum number of evaluations to be 2,000 for both EoH and FunSearch and EoH's population size to be 20.

\begin{figure}[t]
     \centering
     \begin{subfigure}[b]{0.47\textwidth}
         \centering
         \includegraphics[width=\textwidth]{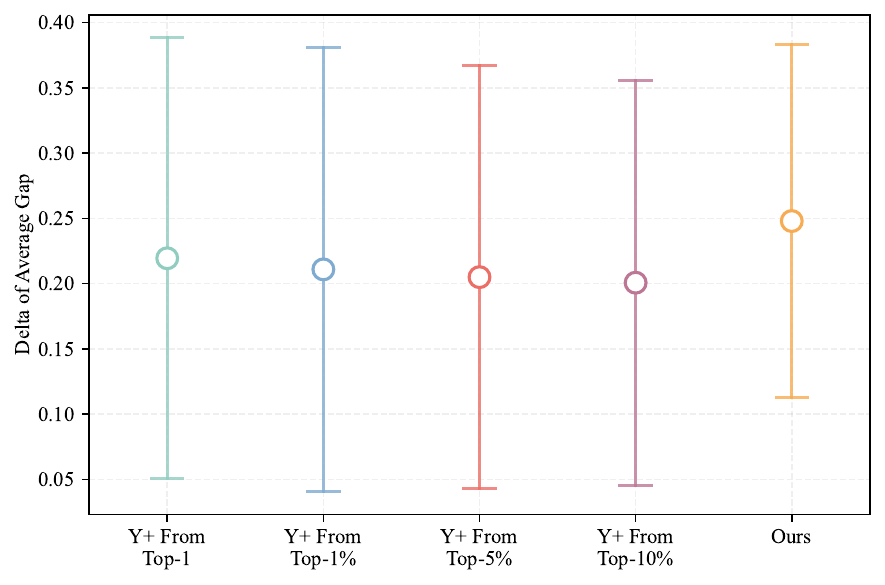}
         \caption{}
         \label{fig:expt1_1_a}
     \end{subfigure}\hfill
     \begin{subfigure}[b]{0.47\textwidth}
         \centering
         \includegraphics[width=\textwidth]{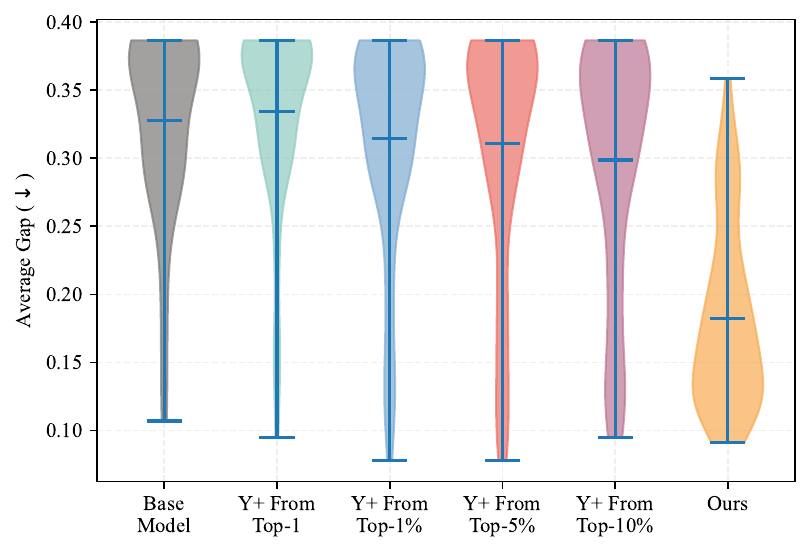}
         \caption{}
         \label{fig:expt1_1_b}
     \end{subfigure}\hfill
    \caption{Comparison on varying preference pair sampling settings. \textbf{(a):} Comparison of the delta value of preference pairs sampled by different methods. The delta value is calculated by the absolute difference of the average gap between $y_+$ and $y_-$ data. The mean delta values of 250 preference pairs are highlighted with circle markers, while the standard deviations are denoted by lines. \textbf{(b):} Violin plot comparison on the performance of LLMs fine-tuned on various preference datasets. Each violin reflects the performance distribution of the top 50 of 1,000 algorithms generated by each LLM. The performance is determined by the average gap to the existing best-known algorithm.}
    \label{fig:expt1_1}
\end{figure}

\vspace{-0.5em}
\subsection{Comparative Evaluation of Sampling Strategies}
In this experiment, we evaluate the efficacy of our proposed sampling strategy against top-k-based sampling strategies. 
% A natural and intuitive method for constructing preference pairs is to define a fitness threshold and sample positive examples ($y_+$) from heuristics above this threshold and negative examples ($y_-$) from those below it. A similar idea is also adopted in prior work such as EvoTune \cite{surina2025algorithm}, where only those pairs in which the $y_+$ exhibits top-tier performance are retained for training. Inspired by this idea, 
We construct four distinct datasets using the top-k-based sampling approach under the following configurations: 1) \textbf{Top-1 Sampling:} The positive samples ($y_+$) are selected as the best-performing heuristic (exhibiting the lowest average gap) in the database, while the negative samples ($y_-$) are randomly chosen from the remaining heuristics. 2) \textbf{Top-k\% Sampling:} The $y_+$ samples are randomly drawn from the top-k\% heuristics, whereas the $y_-$ samples are selected from the rest $(100-k)\%$. We test different parameters $k = 1, 5, 10$ for this strategy.

We evaluate the performance of the Llama-1B model, which has been fine-tuned on ASP. To quantify the differences between these strategies, we analyze the delta value of preference pairs, defined as the difference in average gap values between $y_+$ and $y_-$ samples within each pair. This metric reflects the relative performance gap between the compared heuristics.

As illustrated in Figure \ref{fig:expt1_1_a}, we visualize the delta value distributions for datasets generated by the five sampling strategies, each comprising 250 preference pairs. The mean delta values (marked with circles) and standard deviations (denoted by lines) reveal that the proposed sampling strategy significantly increases the pairwise distances between preference samples compared to top-k-based methods.

% \paragraph{Zero-shot Performance of Fine-tuned LLMs}
We assess the effectiveness of different sampling methods as follows: Firstly, we randomly sample 1,000 feasible algorithms that have been successfully evaluated on ASP using both the base model and its fine-tuned variants. The prompts used for this random sampling are identical to $x$ in the preference pair. Subsequently, we identify and analyze the top 50 algorithms from this pool of 1,000, plotting the distribution of their average gap values. As illustrated in Figure \ref{fig:expt1_1_b}, fine-tuning with our proposed sampling strategy markedly enhances the model's capability in designing algorithms. This improvement is quantitatively demonstrated by a significantly reduced mean average gap among the top-50 algorithms when compared to those designed by the base model and other sampling methods.
% We assess the performance of the \textit{Llama-3.2-1B-Instruct} model fine-tuned on these datasets for the ASP using the following protocol:
% \begin{enumerate}
%     \item First, we randomly sample 1,000 feasible algorithms (successfully evaluated) on ASP using the base model and fine-tuned variants. The prompt we used in random sampling is identical to $x$ in the preference pair. 
%     \item Then, we filter the top 50 algorithms among 1,000 feasible algorithms, and we plot the distribution of the average gap values for these 50 algorithms.
% \end{enumerate}
% Figure \ref{fig:expt1_1_b} demonstrates that fine-tuning with the proposed sampling strategy substantially improves the LLM's algorithm design capability, as evidenced by a significantly lower mean average gap in the top-50 algorithms compared to the base model and other fine-tuned variants.

\subsection{Performance Improvement via Fine-tuning}\label{sec:expt1}
% In this section, we first systematically explore strategies for building effective DPO datasets. Then we empirically study whether fine-tuning can enable a smaller LLM to approach the performance of larger models. 

\paragraph{Random Sampling} \label{sec:zero_shot_performance}
In this experiment, we investigate whether a fine-tuned smaller LLM can match larger LLMs in terms of algorithm design capabilities. Specifically, we fine-tune Pangu-1B and Llama-1B on datasets of 2,000 and 5,000 preference pairs and compare them with their base models. On ASP, we further include Pangu-7B and Llama-8B as larger auxiliary comparison models.

\begin{figure}[t]
     \centering
     \begin{subfigure}[b]{0.6\textwidth}
         \centering
         \includegraphics[width=\textwidth]{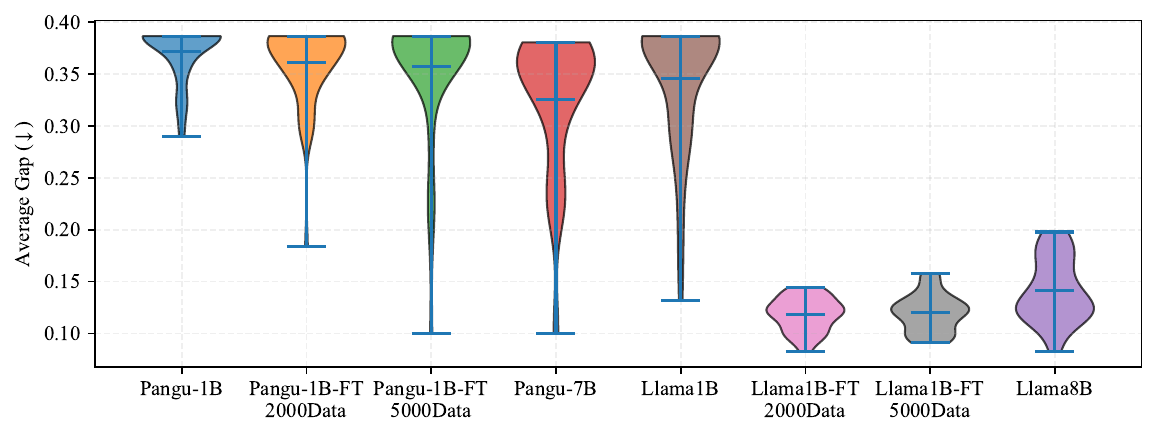}
        %  \caption{Performance on Admissible Set}
     \end{subfigure}
    \caption{Violin plot comparison on the performance of fine-tuned LLMs and base model on ASP. Each violin reflects the performance distribution of the top 50 of 1,000 algorithms generated by each LLM. The performance is determined by the average gap to the existing best-known algorithm, with lower being better.}
    \label{fig:expt1_2}
\end{figure}

Figure \ref{fig:expt1_2} compares the performance of the base models, the 1B models fine-tuned on two dataset sizes, and the larger auxiliary comparison models. Similar to the previous experiment, we randomly sample 1,000 feasible algorithms and plot the distribution of the average gap of the top 50 algorithms, with lower being better. It can be concluded from the results that: i) Fine-tuning consistently improves the two 1B models on ASP. ii) Increasing the dataset size yields marginal but consistent improvements. iii) The fine-tuned Llama-1B achieves competitive performance with Llama-8B, highlighting the effectiveness of our approach. iv) Although fine-tuning also improves Pangu-1B, its average gap remains higher than that of the larger Pangu-7B model.

\begin{figure}[t]
     \centering
     \begin{subfigure}[b]{0.47\textwidth}
         \centering
         \includegraphics[width=\textwidth]{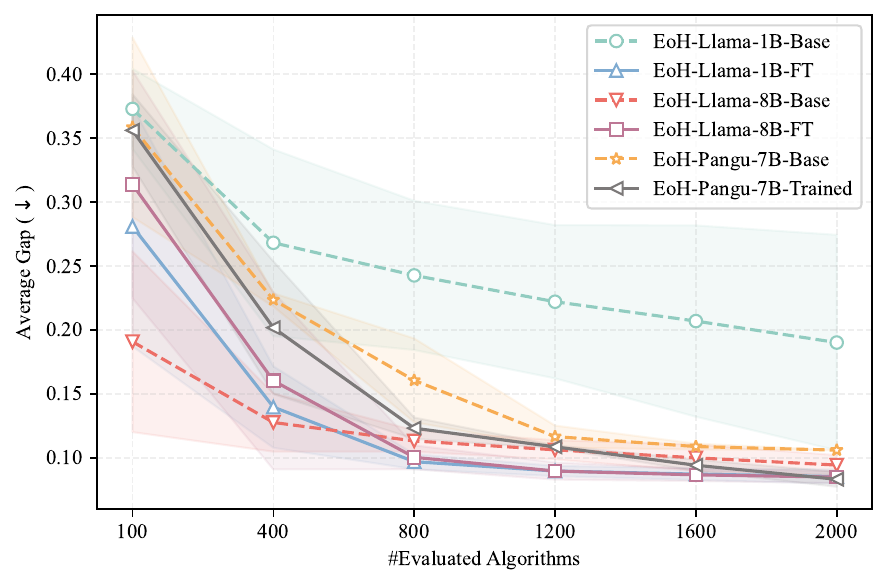}
         \caption{EoH}
     \end{subfigure}\hfill
     \begin{subfigure}[b]{0.47\textwidth}
         \centering
         \includegraphics[width=\textwidth]{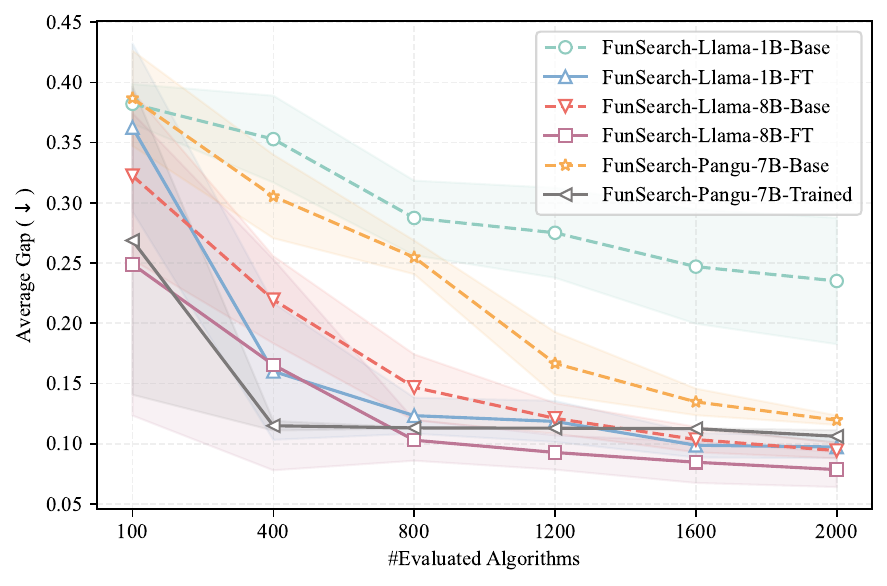}
         \caption{FunSearch}
         \label{fig:ll}
     \end{subfigure}\hfill
    \caption{Convergence curve comparison on the performance of top-5 algorithms generated by various LLMs and search methods on ASP. The performance is determined by the average gap to the best-known algorithm, with lower being better. The mean performance averaged over three independent runs is denoted by markers, while the standard deviations are demonstrated by the shaded area.}
    \label{fig:admi_search}
\end{figure}

\begin{table}[h]
    \caption{Performance comparison on ASP for the top-1 and top-10 heuristics. The performance is determined by the average gap to the best-known optimum (\%). The mean and standard deviation aggregated over three independent runs are reported, with lower being better.}
    \centering
    \resizebox{0.9\textwidth}{!}{
    \begin{tabular}{ll|cc|cc|cc}
    \toprule
     & & Llama-1B & Llama-1B-FT & Llama-8B & Llama-8B-FT & Pangu-7B & Pangu-7B-FT \\
     \midrule
    \multirow{2}{*}{FunSearch} & Top-1 & $19.48_{\pm 4.31}$ & \cellcolor{gray!25}$9.19_{\pm 0.28}$ & $8.46_{\pm 0.46}$ & \cellcolor{gray!25}$7.43_{\pm 1.22}$ & $11.00_{\pm 1.23}$ & \cellcolor{gray!25}$10.94_{\pm 0.49}$ \\
& Top-10 & $24.22_{\pm 5.03}$ & \cellcolor{gray!25}$9.88_{\pm 0.97}$ & $9.88_{\pm 0.84}$ & \cellcolor{gray!25}$8.09_{\pm 1.5}$ & $13.35_{\pm 1.87}$ & \cellcolor{gray!25}$11.19_{\pm 0.34}$ \\

    \midrule

    \multirow{2}{*}{EoH} & Top-1 & $17.02_{\pm 8.88}$ & \cellcolor{gray!25}$8.52_{\pm 0.56}$ & $8.99_{\pm 0.62}$ & \cellcolor{gray!25}$8.19_{\pm 0.53}$ & $8.81_{\pm 0.23}$ & \cellcolor{gray!25}$8.68_{\pm 0.59}$ \\
& Top-10 & $19.49_{\pm 8.34}$ & \cellcolor{gray!25}$8.58_{\pm 0.52}$ & $9.83_{\pm 0.85}$ & \cellcolor{gray!25}$8.69_{\pm 0.46}$ & $9.69_{\pm 0.72}$ & \cellcolor{gray!25}$9.66_{\pm 0.79}$ \\

    \bottomrule
    \end{tabular}
    }
    \label{tab:admi_search}
\end{table}

\begin{table}[h]
    \caption{Performance comparison on CVRP for the top-1 and top-10 heuristics. The performance is determined by the average gap to the best-known optimum (\%). The mean and standard deviation aggregated over three independent runs are reported, with lower being better.}
    \centering
    \resizebox{0.7\textwidth}{!}{
    \begin{tabular}{ll|cc|cc}
    \toprule
     & & Llama-1B & Llama-1B-FT & Llama-8B & Llama-8B-FT \\
     \midrule
    \multirow{2}{*}{FunSearch} & Top-1 & $35.81_{\pm 0.90}$ & \cellcolor{gray!25}$30.26_{\pm 2.02}$ & $30.23_{\pm 1.73}$ & \cellcolor{gray!25}$27.38_{\pm 2.06}$ \\
& Top-10 & $37.81_{\pm 0.91}$ & \cellcolor{gray!25}$33.27_{\pm 4.23}$ & $31.91_{\pm 2.89}$ & \cellcolor{gray!25}$27.98_{\pm 2.38}$ \\

    \midrule

    \multirow{2}{*}{EoH} & Top-1 & $36.98_{\pm 0.15}$ & \cellcolor{gray!25}$31.61_{\pm 4.05}$ & $28.45_{\pm 3.21}$ & \cellcolor{gray!25}$27.06_{\pm 2.7}$ \\
& Top-10 & $37.02_{\pm 0.11}$ & \cellcolor{gray!25}$34.64_{\pm 3.51}$ & $30.22_{\pm 3.29}$ & \cellcolor{gray!25}$27.38_{\pm 2.5}$ \\

    \bottomrule
    \end{tabular}
    }
    \label{tab:cvrp_search}
\end{table}

% \subsection{Coupling Fine-tuned LLMs with Search}
\paragraph{Iterative Search} \label{sec:search_performance}
We couple the fine-tuned LLM with two state-of-the-art LLM-driven AAD methods, FunSearch and EoH. We initialize all compared methods with the respective seed algorithm on each problem. We set the maximum number of evaluated programs to 2,000. The maximum evaluation time for each heuristic is restricted to 30 seconds to eliminate inefficient and harmful heuristics, such as infinite loops. We perform three independent runs for each method to account for experimental biases.

We evaluate Llama-1B and Llama-8B in this study. For the ASP problem, we additionally include Pangu-7B and its fine-tuned variant as auxiliary comparison models. For each task, we generate a dataset of 2,000 preference pairs using our proposed sampling strategy and fine-tune the corresponding models considered in that task.
After that, the fine-tuned LLMs, as well as the base models, are utilized to generate algorithms under the guidance of search methods. 

\begin{figure}[t]
     \centering
     \begin{subfigure}[b]{0.47\textwidth}
         \centering
         \includegraphics[width=\textwidth]{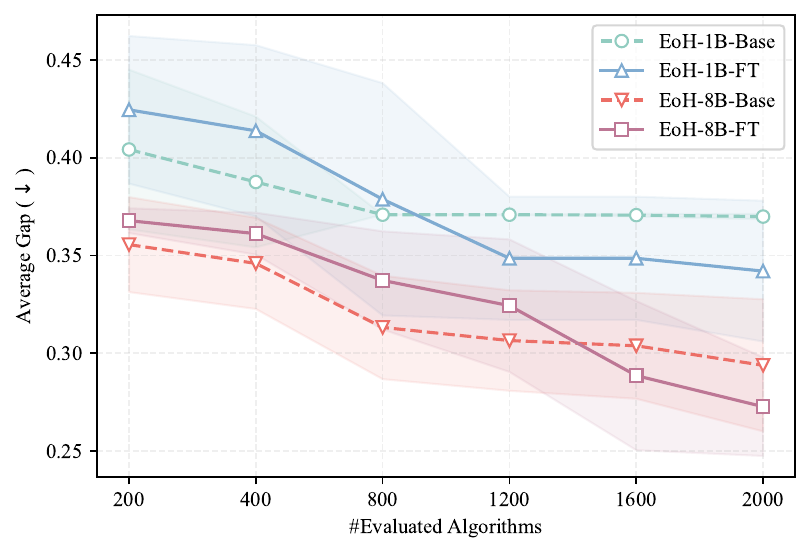}
         \caption{EoH}
     \end{subfigure}\hfill
     \begin{subfigure}[b]{0.47\textwidth}
         \centering
         \includegraphics[width=\textwidth]{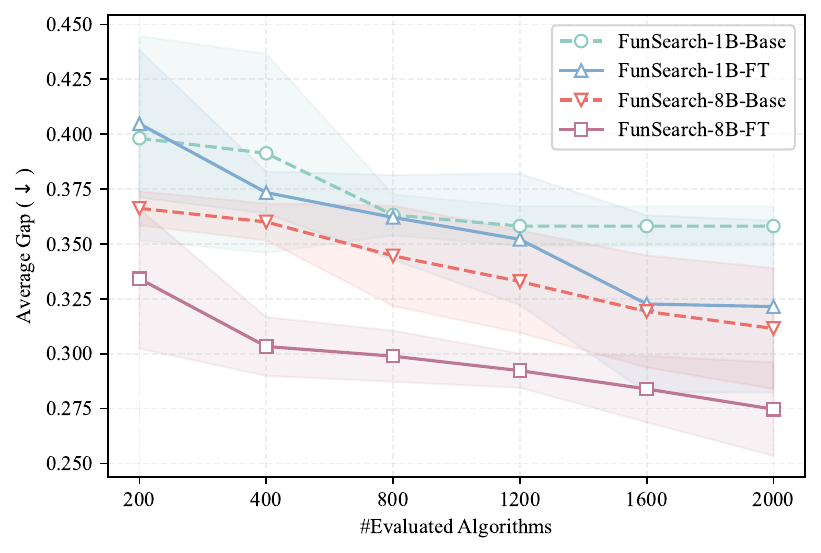}
         \caption{FunSearch}
         \label{fig:cvrp_eoh}
     \end{subfigure}\hfill
    \caption{Convergence curve comparison on the performance of top-5 algorithms generated by various LLMs and search methods on CVRP problem. The performance is determined by the average gap to the existing best-known algorithm. The mean performance averaged over three independent runs is denoted by markers, while the standard deviations are demonstrated by the shaded area.}
    \label{fig:cvrp_search}
\end{figure}

Figures \ref{fig:admi_search} (ASP problem) and \ref{fig:cvrp_search} (CVRP problem) present the convergence curves of the performance of top-5 algorithms, with markers indicating mean performance across three runs and shaded regions denoting standard deviations. Complementary results for the performance of top-1 and top-10 algorithms are listed in Tables \ref{tab:admi_search} and \ref{tab:cvrp_search}. Results show that: i) The fine-tuned LLMs consistently outperform base models, achieving faster convergence and smaller optimality gaps.
ii) The 1B LLMs can approach the performance of 8B LLMs in most cases, with the exception of EoH on the ASP, where the 1B LLM is noticeably inferior to the base 8B LLM.

These results demonstrate that preference learning enhances LLMs’ algorithm design capabilities, which in turn elevate search performance. This underscores the importance of integrating search strategies with fine-tuned LLMs.

\begin{figure}[t]
     \centering
     \begin{subfigure}[b]{0.47\textwidth}
         \centering
         \includegraphics[width=\textwidth]{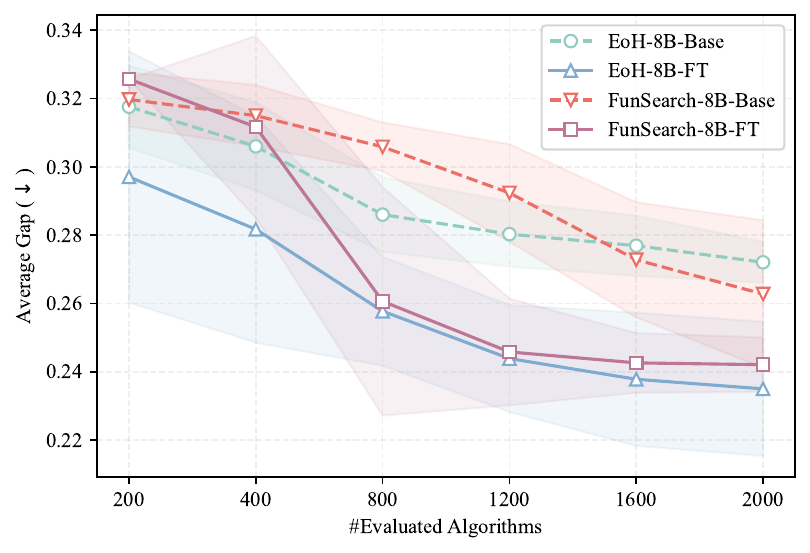}
         \caption{CVRP-100 Problem}
     \end{subfigure}\hfill
     \begin{subfigure}[b]{0.47\textwidth}
         \centering
         \includegraphics[width=\textwidth]{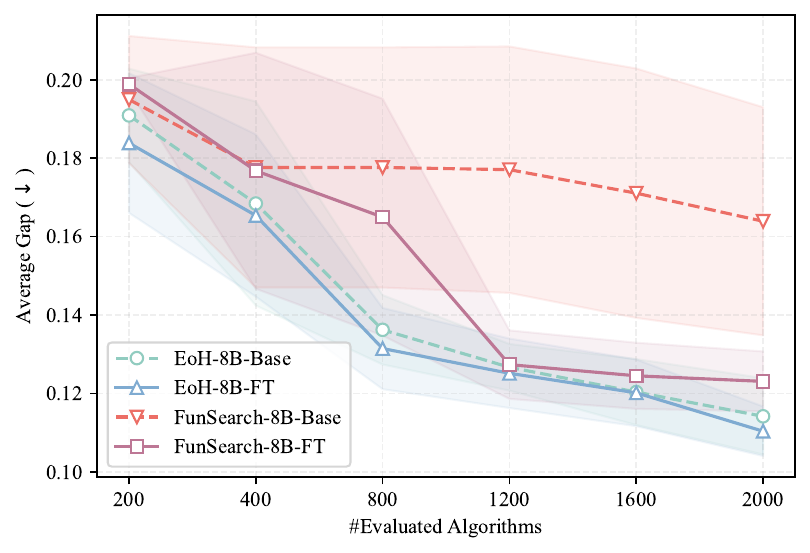}
         \caption{TSP-50 Problem}
     \end{subfigure}\hfill
    \caption{Convergence curve comparison on the performance of top-5 algorithms generated by the base model and LLMs fine-tuned on the CVRP-50 problem. The performance is determined by the average gap to the existing best-known algorithm, with lower being better. The mean performance averaged over three independent runs is denoted by markers, while the standard deviations are demonstrated by the shaded area.}
    \label{fig:convergence_ood}
\end{figure}

% \subsection{Grid-based Training Sample Construction}

\begin{table}[t]
    \caption{Performance comparison on the top-1 and top-10 heuristics. The performance is determined by the average gap to the best-known optimum (\%). The mean and standard deviation aggregated over three independent runs are reported, with lower being better.}~\label{table:ood}
    \centering
    \resizebox{0.7\textwidth}{!}{
    \begin{tabular}{ll|cc|cc}
    \toprule
     \multirow{2}{*}{} & & \multicolumn{2}{c|}{Performance on CVRP-100} & \multicolumn{2}{c}{Performance on TSP-50}  \\
     \cmidrule(lr){3-4}\cmidrule(lr){5-6}
     & & Top-1 & Top-10 &  Top-1 & Top-10  \\
     \midrule
     
    \multirow{2}{*}{FunSearch} & Llama-8B & $26.27_{\pm 2.16}$ & $26.74_{\pm 1.95}$ & $15.53_{\pm 2.51}$ & $16.88_{\pm 2.98}$ \\
& \cellcolor{gray!25}Llama-8B-FT & \cellcolor{gray!25}$24.1_{\pm 0.87}$ & \cellcolor{gray!25}$24.26_{\pm 0.84}$ & \cellcolor{gray!25}$11.85_{\pm 0.37}$ & \cellcolor{gray!25}$12.52_{\pm 0.73}$ \\

    \midrule
    
    \multirow{2}{*}{EoH} & Llama-8B & $26.88_{\pm 0.42}$ & $27.47_{\pm 0.69}$ & $10.87_{\pm 0.84}$ & $11.76_{\pm 1.03}$ \\
& \cellcolor{gray!25}Llama-8B-FT & \cellcolor{gray!25}$23.34_{\pm 2.01}$ & \cellcolor{gray!25}$23.62_{\pm 1.95}$ & \cellcolor{gray!25}$10.3_{\pm 0.74}$ & \cellcolor{gray!25}$11.57_{\pm 0.83}$ \\

    \bottomrule
    \end{tabular}
    }
    \label{tab:my_label}
\end{table}

\subsection{Generalization of Fine-tuned LLMs} \label{sec:generalized_performance}

In this section, we investigate the capacity of online fine-tuned LLMs to enhance algorithm search on new algorithm design tasks. We employ the Llama-8B model. The model has been fine-tuned on the CVRP with instance size 50, and we adopt it on two new tasks without any further adaptation. These two settings represent different levels of generalization: 1) The same task under a different distribution. 2) A related task with a different problem description.

\paragraph{Same Task with Different Distribution}
We evaluated the performance of the fine-tuned Llama-8B model on a variant of the CVRP. Originally fine-tuned for CVRP instances with 50 nodes, the model is now tested on larger instances containing 100 nodes. The coordinates for these nodes are randomly generated within the [0,1] interval, and the vehicle capacity has been increased to 50. This setup introduces variations in both the number of nodes and the capacities, while the task description remains consistent. Results in Figure~\ref{fig:convergence_ood} and Table~\ref{table:ood} demonstrate that the fine-tuned LLMs are effective when we change the task settings. The average results clearly outperform base LLMs on both EoH and FunSearch.  

% \subsubsection{Related Task with Similar Description}

% The model is evaluated on the Vehicle Routing Problem with Open Routes (OVRP), where vehicles do not return to depots after serving all customers. OVRP shares similarities with CVRP but relaxes the depot return constraint, providing a test of the model's adaptability to related but distinct tasks.

\paragraph{Related Task with Different Description}
The fine-tuned Llama-8B is also tested on the Traveling Salesman Problem (TSP), which, while related to CVRP, differs significantly in terms of problem description and attributes. The results in Figure~\ref{fig:convergence_ood} demonstrate that even on a different task, fine-tuned LLMs can still improve automated algorithm search. However, the improvement is less pronounced on EoH, likely because EoH already converges efficiently with the base model.

\section{Conclusion}

This paper presents a preliminary study on the necessity and effectiveness of fine-tuning an LLM tailored to the algorithm design task. We adopt DPO and introduce a diverse-aware rank-based sampling strategy, which balances training data diversity and quality for effective finetuning on algorithm design tasks. Our experiments on three tasks demonstrate the effectiveness of the fine-tuned LLM across different algorithm design scenarios, including: algorithm design with LLM-based random sampling, algorithm design with LLM-based iterative search, and generalizing to related algorithm design tasks. Notably, Llama-1B trained with our method matches the performance of Llama-8B.

% Future research can further extend this work to an online learning scenario, where the model iteratively refines its performance based on dynamically generated feedback or solutions.

% This study preliminarily demonstrates that fine-tuning LLMs for automated algorithm design tasks significantly improves their performance and generalization capabilities in various algorithm design tasks. 
% Our diversity-aware ranked sampling strategy enhances the diversity and quality of training data, enabling LLMs to learn more effectively. 
% Notably, our results show that the effectiveness of fine-tuning on improving LLM's algorithm design performance.
% Furthermore, the ability of these models to generalize across related tasks highlights their utility in addressing diverse algorithmic challenges.
% This research provides a preliminary study of how and to what extent we can effectively fine-tune LLMs to enhance automated algorithm design. 

% Beyond immediate applications, our findings also suggest promising directions for improving the efficiency of LLM fine-tuning. 

\bibliography{references}
\bibliographystyle{iclr2026_conference}

% \appendix
% \section{Appendix}

\end{document}